\documentclass[runningheads,a4paper]{llncs}

\usepackage[utf8]{inputenc}
\usepackage[T1]{fontenc}    % use 8-bit T1 fonts
\usepackage{hyperref} 
\usepackage{graphicx,times,amssymb,float,xcolor}      % hyperlinks
\usepackage{url}           % simple URL typesetting
\usepackage{epstopdf,epsfig}    
\usepackage{booktabs}       % professional-quality tables
\usepackage{amsfonts}      % compact symbols for 1/2, etc.
\usepackage{microtype}      % microtypography
\usepackage{bm}
\usepackage{caption}
\DeclareCaptionType{noticebox}
\usepackage{subcaption}
\usepackage{pdfpages}
\usepackage{amsmath}
\usepackage{cleveref}
\usepackage[linesnumbered,algoruled,boxed,lined]{algorithm2e}
\usepackage{algorithmic}
\usepackage{tabularx}
\usepackage[export]{adjustbox}

\DeclareMathOperator*{\argmax}{arg\,max}

\DeclareMathOperator*{\argsort}{arg\,sort}

\newcommand{\ie}{i.e.}

\DeclareMathOperator*{\argmin}{arg\,min}
\DeclareCaptionLabelFormat{andtable}{#1~#2  \&  \tablename~\thetable}

\usepackage{url}
\urldef{\mailsa}\path|{alfred.hofmann, ursula.barth, ingrid.haas, frank.holzwarth,|
\urldef{\mailsb}\path|anna.kramer, leonie.kunz, christine.reiss, nicole.sator,|
\urldef{\mailsc}\path|erika.siebert-cole, peter.strasser, lncs}@springer.com|

\newcommand\cincludegraphics[2][]{\raisebox{-0.15\height}{\includegraphics[#1]{#2}}}

\begin{document}
\allowdisplaybreaks
\mainmatter  % start of an individual contribution

\title{Feature Selection for Unsupervised Domain Adaptation using Optimal Transport}
\titlerunning{Feature Selection for Unsupervised Domain Adaptation using Optimal Transport}


\author{L\'eo Gautheron\inst{1\thanks{Most of the work in this paper was carried out when L.G. was affiliated with CREATIS.}} \and Ievgen Redko\inst{2} \and Carole Lartizien \inst{2}}

%Craig Chambers, Kim B. Bruce, and Elisa Bertino}
%

\institute{
    Univ Lyon, UJM-Saint-Etienne, CNRS, Institut d Optique Graduate School\\
    \ \ \ Laboratoire Hubert Curien UMR 5516, F-42023, SAINT-ETIENNE, France\\
   \email{leo.gautheron@univ-st-etienne.fr}
    \and
    Univ Lyon, INSA‐Lyon, Universit\'e Claude Bernard Lyon 1, UJM-Saint Etienne,\\
    CNRS, Inserm, CREATIS UMR 5220, U1206, F-69621, LYON, France\\
    \email{\{name.surname\}@creatis.insa-lyon.fr}\\
    }



% TODO Leo
% - PLOT RUNNING TIME vs SAMPLE SIZE
%  - running time sample selection
%  - running time transport features
% - MENTION THE VALUE OF \lambda 
% - EXPERIMENTS FACES DIGITS VOIRE PAPIER OTDA

%\toctitle{Lecture Notes in Computer Science}
%\tocauthor{Authors' Instructions}
\maketitle
    \begin{abstract}
        %In many real-world applications, it may be desirable to reuse a classifier learned on a benchmark annotated data set in order to apply it further on different yet similar scarcely annotated data set. The area of machine learning dealing with this particular learning scenario, called \textit{domain adaptation}, aims at providing solutions that allow to overcome the discrepancy between the samples used during the training and testing stages. 
        %Domain adaptation is a popular research area in machine learning that studies the problem occurring when data samples used to train and test a classifier do not follow the same probability distribution. 
        In this paper, we propose a new feature selection method for unsupervised domain adaptation based on the emerging \textit{optimal transportation theory}. We build upon a recent theoretical analysis of optimal transport in domain adaptation and show that it can directly suggest a feature selection procedure leveraging the shift between the domains. Based on this, we propose a novel algorithm that aims to sort features by their similarity across the source and target domains, where the order is obtained by analyzing the coupling matrix representing the solution of the proposed optimal transportation problem. We evaluate our method on a well-known benchmark data set and illustrate its capability of selecting correlated features leading to better classification performances. Furthermore, we show that the proposed algorithm can be used as a pre-processing step for existing domain adaptation techniques ensuring an important speed-up in terms of the computational time while maintaining comparable results. Finally, we validate our algorithm on clinical imaging databases for computer-aided diagnosis task with promising results.
    \end{abstract}
    
    \section{Introduction}
    The majority of well-known machine learning algorithms used in real-world applications are built upon the common strategy often known as empirical risk minimization. This strategy suggests that a classifier that minimizes the loss over the observed samples is expected to generalize and thus to perform well on any other sample coming from the same probability distribution. However, this assumption is often violated in practice where a training sample may be different from new unseen data collected afterwards. For instance, one may consider the spam filtering problem. It is quite intuitive to suggest that a given user will be targeted with spam messages depending on its browsing history and that a classifier distinguishing between spam and non spam messages may not be equally efficient for two different users if it does not adapt correctly. In order to tackle this problem, a new learning paradigm called domain adaptation was proposed \cite{Ben-david07analysisof}.
    
    The main goal of domain adaptation is to provide methodological frameworks and algorithms that allow to reuse a classifier learned in one area, usually called \textit{source domain}, in a different yet similar area usually called \textit{target domain}. According to the domain adaptation theory presented in \cite{Ben-david07analysisof,bendavidth}, the efficiency of a given adaptation algorithm depends on its capacity to reduce the discrepancy between the probability distributions of the considered source and target samples and on the existence of a good hypothesis (or classifier) that can minimize both source and target error functions. While finding this optimal hypothesis is a very difficult problem, most domain adaptation algorithms concentrate solely on reducing the discrepancy between two domains based on the observed samples. To this end, several papers \cite{uguroglu11,persello2016kernel,li2016joint,yin2017cross} proposed to solve the domain adaptation problem by addressing it as a feature selection task. Indeed, for the general adaptation scenario, it is reasonable to assume that the shift between source and target domains may be caused by a changing behavior of a subset of features that characterize the data in both domains. In this case, identifying these features can help to reduce the discrepancy between the source and target domains samples and to allow efficient adaptation. 
    
    %According to a survey on domain adaptation \cite{Marg:2011:REV} there are four main classes of approaches that differ in the assumptions they make: instance weighting for covariate shift \cite{Shimodaira2000,Zadrozny:2004:LEC:1015330.1015425,Huang:2006:CSS:2976456.2976532} (matching reweighting instances between domains); self-labeling methods (guessing labels in target domain and adjusting them during the learning procedure); cluster-based learning (attributing the same label to the instances from dense regions); and feature representation learning \cite{Blitzer:2006:DAS:1610075.1610094,Ando:2005:FLP:1046920.1194905,DBLP:conf/aaai/PanKY08} (projecting initial features to a new space in order to find invariant components).
    
    In this paper, we propose a new feature selection algorithm for unsupervised domain adaptation that allows to rank features based on their similarity across the source and target domains. Our key underlying idea is to solve the optimal transportation problem between the marginal distributions of features in the two domains in order to obtain a coupling matrix given by their joint probability distribution. The goal, then, is to use this coupling matrix to identify the most correlated features by analyzing the diagonal of the coupling matrix where higher coupling values indicate strong correlations between the source and target features. We note that contrary to the state-of-the-art methods that proceed by learning a new richer feature representation before identifying the invariant features, our method performs feature selection directly in the input space. This choice leads to more interpretable results and to a better understanding of the adaptation phenomenon as transformed features cannot directly point out to those descriptors that vary between the two domains. Furthermore, the shifted features identified by our method can be eliminated in order to speed-up domain adaptation algorithms whose running time often inherently depends on the dimensionality of the input data. This latter point is quite important as domain adaptation algorithms are often deployed for high-dimensional data arising from computer vision applications. % where eliminating shifted features before using the domain adaptation algorithm can lead a significantly lower running times. 
    Despite its advantages, our method does not aim to outperform the state-of-the-art classification results obtained by powerful feature transformation domain adaptation methods as most of them use a very rich class of mappings to find a new data representation. To this end, the foremost goal of this paper is to show that the proposed feature selection method is not a competitor of the state-of-the-art algorithms but is a complementary tool that provides important benefits both in terms of computational efficiency and better understanding of data. All the results presented in our paper are given in order to illustrate this rather than its superiority in terms of classification accuracy.
    
    The rest of this paper is organized as follows: in Section \ref{sec:soa} we present a short state-of-the-art on feature selection methods in domain adaptation. Section \ref{sec:knowledge} is devoted to the introduction of basic elements related to the optimal transportation theory that are used later. In Section \ref{sec:method}, we show how a theoretical analysis of domain adaptation with optimal transport can be used to derive a new adaptation algorithm based on feature selection. Based on this, we describe the proposed method and the details of its algorithmic implementation. Section \ref{sec:expes} presents experimental evaluations of the proposed method on both a benchmark computer vision data set and a clinical imaging database for computer-aided diagnosis task. Section \ref{sec:conclusions} summarizes our paper by outlining its main contributions and giving the possible future perspectives of this work.
    
    \section{Related works}
    \label{sec:soa}
    %Our work is placed on the intersection of two research fields that are domain adaptation and feature selection. %Feature selection methods can be grouped in three families \cite{guyon2003introduction}: 1) the filters or ranking methods computing an individual score per feature before sorting them, 2) the wrappers assessing the usefulness of subsets of features according to a classifier, 3)  the embedded methods performing the feature selection while training a classifier. Our method belongs to the first family of feature selection methods.
    %A classical method in filter feature selection is mRMR \cite{peng2005feature}. The idea is that a subset of features should both maximize the dependency of the features with the label while minimizing the sum of the dependency between the pairs of features selected. This methods require to compute mutual information, a quantity difficult to approximate in the presence of high-dimensional continuous feature space. In addition, this method was not designed to work in the presence of several domains.
    As classical feature selection methods \cite{guyon2003introduction} are not designed to work well under the assumption of distribution's shift, several methods were specifically proposed in the literature for feature selection in the context of domain adaptation. For instance, in \cite{li2016joint}, the authors search a latent low-dimensional subspace for two domains by jointly preserving the data structure and by selecting a subset of the latent features through a row-sparsity inducing regularization. While being quite effective in terms of classification results, this method, however, has two important drawbacks. First, it does not identify the original features that contribute to efficient adaptation but rather learns their embedding where the distributions' discrepancy is minimized. Second, its optimization procedure makes use of eigenvalue decomposition which has a high computational cost in large-scale applications. Another example of feature selection methods in domain adaptation are \cite{yin2017cross} and \cite{persello2016kernel}. The contribution of the former paper consists in learning a least squares SVM in order to further remove the features that incur the smallest loss of the classification margin between the classes. The method described in the latter paper proposes to solve an optimization problem with two terms: the first term maximizes the relevance between source features and labels using the Hilbert-Schmidt Independence Criterion while the second term minimizes the shift between the domains using kernel embeddings. Contrary to our algorithm, the above mentioned methods are supervised as they both use annotations in the target domain. Finally, the method that is the most similar to ours is the feature selection algorithm for transfer learning presented in \cite{uguroglu11}. In this paper, the authors use a parametric maximum mean discrepancy distance in order to find a weight matrix that allows to identify invariant and shifting features in the original space. As we show in Section \ref{sec:insight}, this method and our contribution are closely related from the theoretical point of view, even though our method remains much more computationally attractive.

    \section{Preliminary knowledge}
    \label{sec:knowledge}
    In this section we give a brief overview of the basic elements related to the optimal transportation theory that are used later. 
    \subsection{Optimal Transport}
    The theory of optimal transport has been introduced by Gaspard Monge in the 18$^{th}$ century and was recently revisited in \cite{Villani2008}. In essence, this theory gives a mathematically founded tool that allows to align arbitrary probability distributions in an optimal way. 
    
    In the discrete case, it can be formalized as follows. Let $\hat{\mu}_S = \frac{1}{N_S}\sum_{i=1}^{N_S}\delta_{x_i^S}$ %\in \mathcal{P}(\Omega_s)$ 
    and $\hat{\mu}_T = \frac{1}{N_T}\sum_{i=1}^{N_T}\delta_{x_i^T}$ %\in \mathcal{P}(\Omega_t)$ 
    be two empirical probability measures defined as uniformly weighted sums of Diracs with mass at locations defined on two point sets $\bm{S} = \{\bm{x}^{S}_i \in \mathbb{R}^d\}_{i=1}^{N_S}$ and $\bm{T} = \{\bm{x}^{T}_i \in \mathbb{R}^d\}_{i=1}^{N_T}$ drawn from arbitrary probability distributions $\mu_S$ and $\mu_T$. The Monge-Kantorovich problem consists in finding a probabilistic coupling $\gamma$ defined as a joint probability distribution over %$\Omega_s \times \Omega_t$ 
    $\bm{S} \times \bm{T}$
    that minimizes the cost of transport w.r.t. a metric $c: \bm{S} \times \bm{T} \rightarrow \mathbb{R}_+$: 
    \begin{align}
    \gamma^* = \argmin_{\gamma \in \Pi(\hat{\mu}_S, \hat{\mu}_T)}\langle \gamma, C\rangle_F,
    \label{equationOT}
    \end{align}
    where $\langle \cdot \text{,} \cdot \rangle_F$ is the Frobenius dot product, $\Pi(\hat{\mu}_S, \hat{\mu}_T) = \lbrace \gamma \in \mathbb{R}^{N_S \times N_T}_+ \vert \gamma \bm{1} = \hat{\mu}_S, \gamma^T \bm{1} = \hat{\mu}_T\rbrace$ is a set of doubly stochastic matrices and $C$ is a dissimilarity matrix, \ie, for $\bm{x}_i^S\in \bm{S}$ and $\bm{x}_j^T\in \bm{T}$, we have $C_{ij} = c(\bm{x}_i^S,\bm{x}_j^T)$ which defines the energy needed to move a probability mass from $\bm{x}_i^S$ to $\bm{x}_j^T$. This problem admits a unique solution $\gamma^*$ and defines a metric on the space of probability measures (called the Wasserstein distance) as follows:
    \begin{equation}
    W(\hat{\mu}_S, \hat{\mu}_T) = \min_{\gamma \in \Pi(\hat{\mu}_S, \hat{\mu}_T)}\langle \gamma, C\rangle_F.
    \end{equation}
    \iffalse
    optimal transport aims to find a coupling matrix $\gamma$ of two distributions defined as a joint distribution on $\bm{S} \times \bm{T}$ with empirical marginals $\hat{\mu}_S$ and $\hat{\mu}_T$ such that for all $x \in \bm{S},y \in \bm{T}$, we minimize the transport cost from $\hat{\mu}_S$ to $\hat{\mu}_T$ with regard to a transportation cost function $c: \bm{S} \times \bm{T} \rightarrow \mathbb{R}^+$, \ie:
    %\begin{equation}\label{equationOT}
    %\gamma^* = \argmin_{\gamma \in \Pi(\hat{\mu}_S, \hat{\mu}_T)}\langle \gamma, C\rangle_F
    %\end{equation}
    where $\langle ., .\rangle_F$ is Frobenius matrix product and $C_{ij} = c(x_i^S,x_j^T)$. The constraint $\Pi(\hat{\mu}_S, \hat{\mu}_T) = \lbrace \gamma \in \mathbb{R}^{N_S \times N_T}_+ \vert \gamma \bm{1} = \hat{\mu}_S, \gamma^T \bm{1} = \hat{\mu}_T\rbrace$ means that, by summing in $\gamma$ the values in one row for each row, we obtain back the vector $\hat{\mu}_S$. Similarly  by summing the values in one column for each column, we obtain back the vector $\hat{\mu}_T$. 
    \fi
    Despite its elegance and simplicity, the formulation of optimal transport given in Equation \ref{equationOT} (abbreviated \textbf{OT}) is a Linear Programming problem that does not scale well because of its computational complexity.
    \subsection{Entropy-regularized optimal transport}
    In order to solve this issue, \cite{cuturi2013} proposed to add the entropic regularization of $\gamma$ to the Equation \ref{equationOT} leading to the following optimization problem:
    \begin{equation}\label{equationOTentropy}
    \gamma^* = \argmin_{\gamma \in \Pi(\hat{\mu}_S, \hat{\mu}_T)}\langle \gamma, C\rangle_F -\frac{1}{\lambda} E(\gamma),
    \end{equation}
    where $E(\gamma) := -\sum_{ij}\gamma_{ij}\log \gamma_{ij}$. The regularized optimal transport (abbreviated \textbf{OT2}) allows the source instances to be transported more or less uniformly to the target instances based on a hyper-parameter $\lambda$ and can be optimized efficiently with the linear time Sinkhorn-Knopp algorithm \cite{knight2008}.

    \subsection{Optimal transport and domain adaptation}
    The use of optimal transport for domain adaptation has been studied for the first time in \cite{courty2014}. %In this paper, the transport cost matrix $C$ is defined as the pairwise squared Euclidean distance between source and target samples. 
    In this work, the authors present a new variant of optimal transport (abbreviated \textbf{OT3}) based on Equation \ref{equationOTentropy} by adding a class regularization $\ell_{\frac{1}{2},1}$:
    \begin{equation}\label{equationOTclass}
    \gamma^* = \argmin_{\gamma \in \Pi(\hat{\mu}_S, \hat{\mu}_T)}\langle \gamma, C\rangle_F -\frac{1}{\lambda} E(\gamma) +\eta \Omega(\gamma),
    \end{equation}
    where the $\Omega(\gamma) = \sum_j \sum_\mathcal{L} \Vert \gamma(I_\mathcal{L}, j)\Vert_1^{1/2}$ term prevents the source instances with different labels to be transported to the same target instance. $I_\mathcal{L}$ represents the list of sample indexes in $\bm{S}$ with label $\mathcal{L}$, and $j$ goes through the sample indexes in $\bm{T}$.
    
    Using the optimal coupling matrix $\gamma^*$ found with Equation \ref{equationOT}, \ref{equationOTentropy} or \ref{equationOTclass}, the authors propose to transport the source samples by solving for each of them:
    \begin{equation}\label{equationTransportSource}
    \hat{x}_i^S=\argmin_{x\in \mathbb{R}^d} \sum_j \gamma^*_{ij}c(x,x_j^T).
    \end{equation}
    In case of the squared Euclidean distance, the closed form solution of this problem can be written as:
    \begin{equation}\label{equationTransportInterpolation}
    \bm{S_a} = \text{diag}\left(\left(\gamma^* \bm{1}\right)^{-1}\right)\gamma^* \bm{T}.
    \end{equation}
    When the marginals $\hat{\mu}_S$ and $\hat{\mu}_T$ are uniform (in practice, this is always the case for us), the Equation \ref{equationTransportInterpolation} is simplified to
    \begin{equation}\label{equationTransportInterpolation2}
    \bm{S_a} = N_s\gamma^*\bm{T}.
    \end{equation}
    With this computation, each source instance is represented as the weighted barycenter of the target instances with which it has the highest values in $\gamma^*$.
    
    For a graphical comparison of the \textbf{OT}, \textbf{OT2} and \textbf{OT3} algorithms, we refer the reader to the Supplementary material.
    
    \section{Proposed approach}
    \label{sec:method}
    In this section we present our main contribution. We start by formally introducing a theoretical result that we use to derive our algorithm.
    
    \subsection{Theoretical insight}
    \label{sec:insight}
    
    From a theoretical point of view, domain adaptation problem is often formalized as follows: we define a domain as a pair consisting of a distribution $\mu_D$ on $\mathcal{X}$ and a labeling function $f_D: \mathcal{X} \rightarrow [0,1]$. A hypothesis class $H$ is a set of functions so that $\forall h \in H, h : \mathcal{X} \rightarrow \lbrace0,1\rbrace$. Using the proposed notations, the definition of an error function can be given as follows.
    \begin{definition}
        Given a convex loss-function $l$, the probability according to the distribution $\mu_D$ that a hypothesis $h \in H$ disagrees with a labeling function $f_D$ (which can also be a hypothesis) is defined as
        $$\epsilon_D (h,f_D) = \mathbb{E}_{x \sim \mu_D} \left[l(h(x),f_D(x))\right].$$
    \end{definition}
    When the source and target error functions are defined w.r.t. $h$ and $f_S$ or $f_T$, we use the shorthand $\epsilon_S (h, f_S) = \epsilon_S (h)$ and $\epsilon_T (h, f_T) = \epsilon_T (h)$. 
    
    The use of optimal transport in domain adaptation was first theoretically analyzed in \cite{DBLP:journals/corr/RedkoHS16}. In this paper, the authors proved that under some mild assumptions imposed on the form of the transport cost function, the source and target error function can be related through the following inequality
    \begin{align}
    \epsilon_T(h) \leq \epsilon_S(h) + W(\mu_S, \mu_T)+\lambda,
    \label{eq:bound_DAOT}
    \end{align}
    where $\lambda$ is the combined error of the ideal hypothesis $h^*$ that minimizes $\epsilon_{S}(h)+\epsilon_T(h)$. This result shows that in order to upper bound the error of a classifier in the target domain, one has to minimize the source error function and the discrepancy between the source and target distributions given by the Wasserstein distance. 
  
     %As in this paper we are interested in the impact that feature selection may have on the efficiency of the adaptation, we would like to show that reducing the discrepancy between the feature spaces is also meaningful and allows to adapt successfully.
        
    Below, we use this result as a starting point in order to develop our approach. To this end, we first notice that the source and target domains can be equivalently seen as $2$-dimensional product spaces $\mathcal{X}_S \times \mathcal{F}_S$ and $\mathcal{X}_T \times \mathcal{F}_T$, where  $\mathcal{X}_S$ (resp. $\mathcal{X}_T$) and $\mathcal{F}_S$ (resp. $\mathcal{F}_T$) denote the source (resp. target) instance and feature spaces. In this case, the probability distributions $\mu_S$ and $\mu_T$ are also product measures supported on $\mathcal{X}_S \times \mathcal{F}_S$ and $\mathcal{X}_T \times \mathcal{F}_T$ and can be written as $\mu_S^X \times \mu_S^f$ and $\mu_T^X \times \mu_T^f$, respectively. Using the results proved in \cite{Talagrand95IHES} for concentration of measures in product spaces, we can upper bound the Wasserstein distance between $\mu_S$ and $\mu_T$ as follows:
    \begin{align*}
    W(\mu_S, \mu_T) \leq W(\mu_S^f, \mu_T^f) + \int_{\mathcal{F}_S} W(\mu_S^X\vert \mu_S^f, \mu_T^X) d\mu_S^f.
    \end{align*}
    Note that in this inequality, measures $\mu_S^f$ (resp. $\mu_T^f$) and $\mu_S^X$ (resp. $\mu_T^X$) can be used interchangeably. We can see that the first term in the right-hand side stands for the Wasserstein distance between the measures defined on the feature spaces while the second term is the expectation of the Wasserstein distance between the source instances measure conditionally on the source features measure  $\mu_S^X\vert \mu_S^f$ and the target instance measure $\mu_T^X$. %This latter term is quite interesting as it can interpreted as a discrepancy between the source and target instance measures taken with respect to a distribution of source features. 
    Now, by plugging it into the learning bound proposed in Equation \ref{eq:bound_DAOT}, we obtain
    \begin{align*}
    \epsilon_T(h) &\leq \epsilon_S(h) + W(\mu_S^f, \mu_T^f)+\int_{\mathcal{F}_S} W(\mu_S^X\vert \mu_S^f, \mu_T^X) d\mu_S^f + \lambda.
    \end{align*}
    This inequality shows that when one considers probability measures over a product space of instances and features spaces, successful adaptation necessitates the minimization of the discrepancy between features distributions $\hat{\mu}_S^f$, $\hat{\mu}_T^f$ as well as that of the instances distributions $\mu_S^X$, $\mu_T^X$ conditionally on the source features measure $\mu_S^f$. Thus, it naturally leads to a two-stage procedure where the first goal is to reduce the discrepancy between the features sets of the two domains while the second is to apply an appropriate domain adaptation algorithm between their instances described by an optimal set of features obtained at the first stage. 
    
    In what follows, we introduce our method based on the idea of finding a coupling that aligns the distributions of features across the source and target domains. As suggested by the obtained bound, the selected features minimizing the $W(\mu_S^f, \mu_T^f)$ can be used then by a domain adaptation algorithm applied to the source and target samples of a reduced dimensionality. We also note that the Wasserstein distance here can be replaced, in practice, by the popular maximum mean discrepancy distance \cite{gretton2006kernel} often used in domain adaptation as both of them belong to a larger class of integral probability metrics defined over different functional classes. In this case, the feature selection algorithm proposed in \cite{uguroglu11}\footnote{Unfortunately, we were unable to use this method as a baseline in our experiments due to the lack of implementation details in their paper and the absence of a publicly available code.} also indirectly minimizes the discrepancy between the marginals $\mu_S^f$ and  $\mu_T^f$. Nevertheless, the computational complexity of the proposed optimization procedure is polynomial thus making its use prohibitive in real-world applications.
    
    %As we show in the following section, using only our method leads to an improved performance over the ``no adaptation setting" but the best results are obtained when it is combined with an actual adaptation algorithm.
    
    \iffalse
    Furthermore, in case where $\mu_S^X$ and $\mu_S^f$ are independent, this simplifies to
    \begin{align}
    W_1(\mu_S, \mu_T) \leq W_1(\mu_S^f, \mu_T^f) + W_1(\mu_S^X, \mu_T^X).
    \end{align}
    
    We can now assume that source and target product measures have finite square-exponential moments, \ie, for some $\alpha >0$ $\mathbb{E}_{x \sim \mu_S}e^{\alpha\Vert x\Vert^2}<\infty$ and $\mathbb{E}_{x \sim \mu_T}e^{\alpha\Vert x\Vert^2}<\infty$, respectively. In this case, by the Talagrand's inequality we have that
    \begin{align}
    W_1(\mu_S,\mu_T) \leq \sqrt{\psi \text{K}(\mu_S\vert \mu_T)}
    \end{align}
    for some $\psi >0$. Here $\text{KL}(\mu_S\vert \mu_T) = \int_\Omega \log \left( \frac{d\mu_S}{d\mu_T}\right)d\mu_S$ stands for the Kullback-Leibler divergence. 
    \fi
    
    \subsection{Problem setup}
    Until now the optimal transport was used in order to align empirical measures $\hat{\mu}_S$ and $\hat{\mu}_T$ defined based on the observable samples $\bm{S} \in \mathbb{R}^{N_S \times d}$ and $\bm{T} \in \mathbb{R}^{N_T \times d}$. The interpolation step performed using Equation \ref{equationTransportInterpolation2}  aims at re-weighting the source instances so that their distribution matches the one of the target samples. The geometric interpretation is that, to minimize the divergence between $\mu_S$ and $\mu_T$, we can associate the source samples with the target samples with which they have the highest coupling values.
    
    As mentioned in the previous section, the idea of our method is to go from the sample space to the feature space. To this end, we now consider that $\bm{S}$ and $\bm{T}$ are drawn from $2$-dimensional product spaces $\mathcal{X}_S \times \mathcal{F}_S$ and $\mathcal{X}_T \times \mathcal{F}_T$, where $\mathcal{X}_S, \mathcal{X}_T \subseteq \mathbb{R}^d$
    %are source and target instance spaces so that $\bm{S} \in \mathcal{X}_S$ and $\bm{T} \in \mathcal{X}_T$
    while $\mathcal{F}_S \subseteq \mathbb{R}^{N_S}$ and $\mathcal{F}_T \subseteq \mathbb{R}^{N_T}$.
    %denote their corresponding feature spaces.
    In this case, we can define two empirical probability measures $$\hat{\mu}_{S}^f = \frac{1}{d}\sum_{i=1}^{d}\delta_{f_i^S} \text{ and  } \hat{\mu}_{T}^f = \frac{1}{d}\sum_{i=1}^{d}\delta_{f_i^T}$$ based on the source and target features $\{f^S\}_{i=1}^d \in \mathcal{F}_S$, $\{f^T\}_{i=1}^d \in \mathcal{F}_T$, respectively. Our goal now would be to transport $\hat{\mu}_{S}^f$ to $\hat{\mu}_{T}^f$ by solving the entropic regularized optimal transportation problem given as follows:
    \begin{equation}\label{equationFeatureSelection}
    \gamma^{*f} = \argmin_{\gamma^f \in \Pi(\hat{\mu}_{S}^f, \hat{\mu}_{T}^f)}\langle \gamma^f, C^f \rangle_F -\frac{1}{\lambda} E\left(\gamma^f\right),
    \end{equation}
    where $C^f_{ij} = \Vert f_i^S - f_j^T \Vert_2^2$.
  
    In what follows, we show that the solution of this problem can lead to a principally different domain adaptation method that is based on feature selection approach rather than on the original instance re-weighting one. 
 
    \subsection{Finding a shared feature representation}
    
    At this point one may notice that in order to apply optimal transport between $\hat{\mu}_{S}^f$ and $\hat{\mu}_{T}^f$, it is necessary to calculate the cost matrix $C^f$ which is possible only if the numbers of source and target instances are equal. Furthermore, as source and target features are described by supposedly shifted distributions, aligning them directly using any arbitrary sets of instances may not be appropriate due to the differences in the representation spaces that may exist across the two domains. In order to tackle both of these problems, we propose to find a matching between the sample number $i, \forall i = \{1, \dots, N_S\}$ describing the source features and the sample number $j, \forall j=\{1,\dots,N_T\}$ describing the target features based on the original optimal transportation problem. More formally, based on the solution $\gamma^*$ of the optimization problem given by Equation \ref{equationOT}, we define the optimal subset of target instances $\bm{T_u}$ as:
    \begin{equation}
    \bm{T_u} := \{x_j\in\bm{T} \vert j=\argmax \gamma_{ij}^*, i\in \{1, \dots, N_S\}\}.
    \end{equation}
    This particular choice of the algorithm \textbf{OT} rather than its regularized versions (\textbf{OT2} and \textbf{OT3}) is explained by the fact that we are interested in a sparse matching between the two sets, i.e., the one limiting the spread of mass\footnote{The empirical justification of using the \textbf{OT} algorithm for sample selection is given in the Supplementary material.}.
    
    This process, summarized in Algorithm \ref{algoSampleSelection}\footnote{$\mathtt{zscore(X)}$: for each column of $X$, subtract its mean and divide by its standard deviation}, is a required preliminary step consisting in finding which examples will be used to describe the features in the source and target domains. The selection stage used to obtain $\bm{T_u}$ relies on the intrinsic capacity of the coupling matrix to describe the probability of associating each source instance with each target instance based on their similarity.
    
    \begin{figure}[!t]
        
    %\hspace{-5mm}
    \begin{minipage}[h]{.46\linewidth}
    \vspace{0pt}  
    \begin{algorithm}[H]
        \caption{Sample selection in target domain}
        \SetKwInOut{Input}{Input}\SetKwInOut{Output}{Output}
        \Input{$\bm{S} \in \mathbb{R}^{N_S\times d}$,\\ $\bm{T}\in \mathbb{R}^{N_T\times d}$}
        \Output{$\bm{T_u} \in \mathbb{R}^{N_S\times d}$ - optimal subset of target instances}
        \label{algoSampleSelection}
        \vspace{0.3mm}
        \begin{algorithmic}
            \STATE $\mathtt{\bm{S} = zscore(\bm{S})}$; $\mathtt{\bm{T} = zscore(\bm{T})}$\vspace{0.4mm}
            \STATE $\mathtt{\gamma^*\leftarrow OT(\bm{S},\bm{T})}$\vspace{0.4mm}
            \STATE $\mathtt{\bm{T_u} \leftarrow \{x_j\in\bm{T} \vert j=\underset{i=1, \dots, N_S}{\text{argmax}} \gamma_{ij}^*\}}$\vspace{0.2mm}
        \end{algorithmic}
    \end{algorithm}
    \end{minipage}%
    \hspace{2mm}
    \begin{minipage}[!t]{.47\linewidth}
    \vspace{0pt} 
    \begin{algorithm}[H]
        \caption{Feature ranking for domain adaptation}
        \SetKwInOut{Input}{Input}\SetKwInOut{Output}{Output}
        \Input{$\bm{S} \in \mathbb{R}^{N_S\times d}$,\\
            $\bm{T}\in \mathbb{R}^{N_T\times d}$}
        \Output{List $F$ of $d$ most similar features from $\bm{S}$ and $\bm{T}$}
        \label{algoSelection}
        \begin{algorithmic}
            %\STATE Generate $\bm{S}$ of dim $x\times d$ representing the different classes s.t. $x\leq N_T$ and $x\leq N_T$
            \STATE $\mathtt{\bm{T_u} \leftarrow Algorithm \ref{algoSampleSelection}} (\bm{S}, \bm{T})$
            \STATE $\mathtt{\bm{S}^T = zscore(\bm{S}^T)}$; $\mathtt{\bm{T_u}^T = zscore(\bm{T_u}^T)}$
            \STATE $\mathtt{\gamma^{*f} = OT2(\bm{S}^T, \bm{T_u}^T, \lambda=1)}$
            \STATE $\mathtt{F}$\small{$\mathtt{= argSortDesc(\{\{\gamma^{*f}\}_{ii} \vert i \in[1,d]\})}$}
        \end{algorithmic}
    \end{algorithm}
    \end{minipage}
    
    \end{figure}
    
    \subsection{Feature selection}
    
    Now, we let $\bm{T} := \bm{T_u}$ meaning that in Equation \ref{equationFeatureSelection} the target features are described by the set $\bm{T_u}$ of the sample instances. Note that if $N_S>N_T$, we invert the roles of $\bm{S}$ and $\bm{T}$ in Algorithm \ref{algoSampleSelection} and instead let $\bm{S} := \bm{S_u}$. Furthermore, in a highly imbalanced classification setting, or in the presence of a large number of instances, we advise to first select a subset of source instances by balancing the samples according to their classes before applying Algorithm \ref{algoSampleSelection}. This selection allows to capture a class information from the source domain without needing labeled samples from the target domain, and thus is still unsupervised w.r.t. the target domain.
    
    We now solve the problem given in Equation \ref{equationFeatureSelection} and obtain the optimal coupling $\gamma^{*f} \in \mathbb{R}^{d \times d}$. Similar to what we have done at the sample selection step, we analyze the values of the coupling matrix in order to determine the less shifted features across the two domains. The important difference, however, is that we sort the features by analyzing only the diagonal of the coupling matrix. This peculiarity is explained by the fact that the values on the diagonal correspond to the similarities between the same features in the shared source and target representation space. By transporting the features with the \textbf{OT2} algorithm, each source feature is transported to its nearest target features. Because of this, if a given feature is shifted across the two domains, then its mass will be uniformly spread on the target features so that its mass on the corresponding target feature will be rather small. Similarly, if a feature is similar between the source and target domains, then the majority of the mass of this source feature should be found on its corresponding target feature.
    
    Based on this idea, we propose to construct the ordered list of features $F$, where the feature number $i$ in $F$ is the one having the $i^\text{th}$ highest coupling value on the diagonal of the coupling matrix, \ie:
    \begin{equation}
    F =  \argsort(\{\{\gamma^{*f}\}_{ii} \vert i \in \{1, \dots, d\}).
    \end{equation}
    By varying the parameter $\lambda$ in \textbf{OT2}, we can spread the mass of a source feature more or less uniformly when transporting it to the target features. Even though one may obtain different coupling values for different values of $\lambda$, it does not affect the order of features returned in $F$ allowing us to fix $\lambda=1$ in all empirical evaluations to avoid hyper-parameter tuning.
    
    The pseudo-code given in Algorithm \ref{algoSelection} summarizes our feature selection method. After having obtained the ordered list of features $F$, we can use its $d^* < d$ first features for the classification problem at hand. It is worth noting that the proposed method can be applied as a pre-processing before using any domain adaptation algorithm to discard the features that are completely different across the two domains. On the other hand, it can also be applied in ``no adaptation setting" to select the common features between training and test data.% and apply the learned classifier directly.
    
    \section{Experimental evaluations}
    \label{sec:expes}
    
    In this section, we provide an empirical study of the proposed algorithm based on the benchmark computer vision Office/Caltech data set and on clinical imaging database for computer-aided diagnostic task. Note that the optimal transport algorithms \textbf{OT} from Algorithm \ref{algoSampleSelection}, \textbf{OT2} from Algorithm \ref{algoSelection} and \textbf{OT3} are available in the Python POT library\footnote{\url{https://github.com/rflamary/POT}}, making our method straightforward to implement. Nevertheless, we make the Python implementation and the data used in our experiments (except the medical data set) publicly available\footnote{\url{https://leogautheron.github.io}} for the sake of reproducibility.
    
    \subsection{Experiments on visual domain adaptation data} 
    
    The main assumption of our method is that not all features are equally useful for adapting a classifier from source domain to the target one. This is especially the case for data sets described by features calculated using the Bag-of-Words (BoW) methods, such as, for instance, the features of the Office \cite{Saenko:2010:AVC}/Caltech \cite{Gopalan:2011:DAO} data set. 
    
    \paragraph{Office/Caltech data set} For this data set, the classification task is to assign an image to a class based on its content. It is composed of 4 domains $A$, $C$, $W$ and $D$ containing 958, 1123, 295 and 157 images, respectively belonging each to one of 10 different classes. These domains form 12 domain adaptation pairs.
    
    In what follows, we use three different types of features: (1) SURF features \cite{bay2006} of size 800 constructed using the BoW method; (2) CaffeNet features \cite{jia2014} that are obtained  by feeding the images to a pre-trained neural network based on the prominent AlexNet \cite{krizhevsky2012}; (3) GoogleNet features \cite{szegedy2015}  obtained in the way identical to CaffeNet features using GoogleNet network. In order to obtain CaffeNet and GoogleNet features, these two neural networks were first trained on ImageNet, a large data set containing millions of images distributed across 1000 different classes. We removed their classification layer of size 1000 to use the output of the previous layer, giving 4096 features for CaffeNet and 1024 features for GoogleNet. Note that we downloaded the pre-trained networks from the Caffe website \cite{jia2014} before using them to extract the features on our images, and this without doing any fine-tuning or any other modification of the networks apart from removing their last layer.

    The experimental protocol used to evaluate the proposed method is based on the one presented in \cite{courty2014}. For each adaptation pair $\bm{S}\rightarrow \bm{T}$, we randomly sample 20 images per class (8 if $\bm{S}$ is $D$). This gives us 200 images (resp. 80) for $\bm{S}$. All images from $\bm{T}$ are considered. We then apply Algorithm \ref{algoSelection} with $\bm{S}$ and $\bm{T}$ to obtain the ordered list of features $F$. For an increasing number of features $d$, we use the $d$ first features of $F$ to, first adapt $\bm{S}$ to $\bm{T}$, and then use a 1-nearest neighbor classifier with the source adapted data as training set to compute the classification accuracy on the target data. We repeat this 19 times and report mean accuracies for each pair. 
    
    \setlength{\tabcolsep}{1mm}
    \begin{table}[!t]
    \centering
    \begin{tabularx}{\textwidth}{>{\centering\arraybackslash} p{0.47\linewidth}>{\centering\arraybackslash} p{0.5\linewidth}}
     \small{
        \begin{tabular}{c c c c}
        \toprule
        DA pairs & $\searrow 512$ & $\nearrow 512$ & $4096$\\
        \midrule
        A$\rightarrow$C & \textbf{74.9$\pm$2.0} & 29.8$\pm$2.4 & 71.7$\pm$3.5\\
        A$\rightarrow$D & \textbf{78.8$\pm$3.5} & 20.4$\pm$2.8 & 76.0$\pm$3.5\\
        A$\rightarrow$W & \textbf{77.6$\pm$1.9} & 20.2$\pm$3.5 & 66.0$\pm$4.6\\
        C$\rightarrow$A & \textbf{83.7$\pm$1.8} & 38.7$\pm$4.5 & 82.1$\pm$2.2\\
        C$\rightarrow$D & \textbf{76.2$\pm$3.6} & 24.1$\pm$3.4 & 74.2$\pm$4.9\\
        C$\rightarrow$W & \textbf{75.4$\pm$3.5} & 20.3$\pm$3.2 & 70.3$\pm$5.3\\
        D$\rightarrow$A & \textbf{75.4$\pm$2.1} & 20.8$\pm$3.8 & 68.7$\pm$2.9\\
        D$\rightarrow$C & 65.0$\pm$2.6 & 21.5$\pm$2.5 & \textbf{66.6$\pm$1.8}\\
        D$\rightarrow$W & \textbf{92.6$\pm$2.0} & 32.8$\pm$5.1 & 91.9$\pm$1.9\\
        W$\rightarrow$A & \textbf{81.5$\pm$1.2} & 18.8$\pm$2.4 & 68.3$\pm$3.0\\
        W$\rightarrow$C & \textbf{72.2$\pm$1.1} & 23.4$\pm$2.1 & 61.2$\pm$2.1\\
        W$\rightarrow$D & \textbf{96.5$\pm$1.5} & 49.7$\pm$3.2 & 96.3$\pm$1.0\\
        \midrule
        Mean & \textbf{79.2$\pm$2.2} & 26.7$\pm$3.3 & 74.4$\pm$3.0\\
        \bottomrule
        \end{tabular}}
        & {\centering
        \cincludegraphics[width=0.8\linewidth,valign=m, height = 2.5cm]{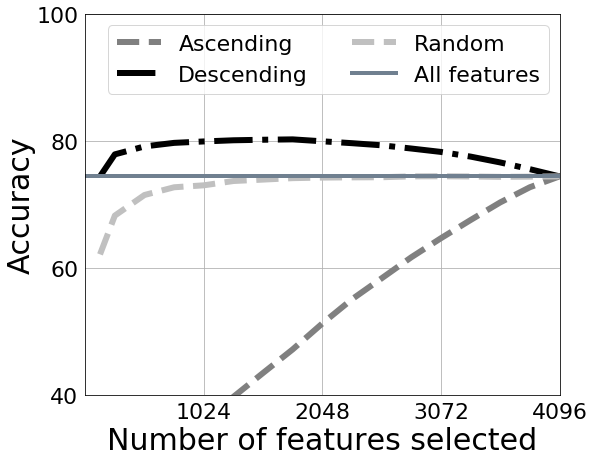}
          } \\
          \vspace{-0.3cm}
          \captionof{table}{\label{officeArrayExp1}} & \vspace{-0.3cm} \captionof{figure}{\label{officeFigExp1}}
            \end{tabularx}
            % \vspace{-0.5cm}
            \captionof*{table}{Results for CaffeNet features. Table \ref{officeArrayExp1} presents mean $\pm$ standard deviation of recognition accuracy with no adaptation. Here, $\searrow X$ (resp. $\nearrow$) indicates the use of the $X$ first features sorted by decreasing (resp. ascending) similarity computed with Algorithm \ref{algoSelection}. In Figure \ref{officeFigExp1}, we plot the mean accuracies from Table \ref{officeArrayExp1}. Our method corresponds to the `Descending' curve consisting in selecting the features ordered by decreasing similarity between source and target domains.}
            \renewcommand*{\arraystretch}{1.05}
	        \setlength{\tabcolsep}{5mm}
            \begin{tabular}{r c c c}
            \toprule
            \#features & SURF & CaffeNet & GoogleNet\\
            \midrule
            $\searrow d/32$ & 21.3$\pm$2.4 & 74.4$\pm$2.9 & 80.0$\pm$2.6\\
            $\nearrow d/32$ & 12.7$\pm$2.0 & 20.6$\pm$3.0 & 24.2$\pm$3.3\\
            \midrule
            $\searrow d/8$ & 25.7$\pm$2.6 & 79.2$\pm$2.2 & 86.9$\pm$1.8\\
            $\nearrow d/8$ & 14.0$\pm$2.2 & 26.7$\pm$3.3 & 48.1$\pm$3.9\\
            \midrule
            $\searrow d/2$ & 29.9$\pm$2.5 & 80.0$\pm$2.2 & 88.1$\pm$1.8\\
            $\nearrow d/2$ & 16.2$\pm$2.5 & 51.3$\pm$4.4 & 77.2$\pm$2.6\\
            \midrule
            d & 27.9$\pm$2.2 & 74.4$\pm$3.0 & 86.8$\pm$1.8\\
            \bottomrule
            \end{tabular}\\
            \caption{\label{officeArrayEx3}Mean accuracies over the 12 DA pairs without applying adaptation using 3 different type of features: SURF (d=800), CaffeNet (d=4096) and GoogleNet (d=1024).}
    \end{table}
    
    \paragraph{Classification results} The classification results for CaffeNet features\footnote{Due to the space limitations, we present the same detailed results for GoogleNet and SURF features in the Supplementary material.} are given in Table \ref{officeArrayExp1}. From this table, we see that by selecting 512 features %out of 4096 
    having the highest similarity between the source and target domains, we obtain a mean accuracy of 79.2\% across the 12 adaptation pairs compared to 74.4\% accuracy obtained using all 4096 features. This behaviour is further confirmed by Figure \ref{officeFigExp1} that illustrates the obtained classification results for a number of features varying between 128 and 4096. %This figure clearly shows that our method identifies features that help to adapt efficiently regardless their number. 
    %In addition, we also show the obtained accuracy results by selecting the features randomly. 
    From it, we also observe that our method outperforms random feature selection while selecting the least similar features gives worse performances in all cases.  
    
    %On the other hand, selecting the 512 features having the lowest similarity gives an accuracy of 27.0\% which confirms the capacity of our method to identify both helpful and harmful features in the sense of their impact on the classification performances. 
    
	The general comparison for CaffeNet features, SURF and GoogleNet features is given in Table \ref{officeArrayEx3}. As before, we observe an important difference between taking the first most similar and dissimilar features across the two domains and note that better performances are obtained by taking a reduced number of features. Another noticeable point is that the performances of SURF features are far behind CaffeNet features, itself slightly worse than GoogleNet features. Even by taking a small number of $1024/32=32$ GoogleNet features, we obtain a mean accuracy of 80.0\% which is at least as good as all the other configurations using SURF and CaffeNet features. To summarize, the presented results clearly show that the order of features returned by our method is directly correlated with their adaptation capacities.
    
    \begin{figure}[!t]
    \centering
    \begin{minipage}{0.45\linewidth}
      \includegraphics[width=\columnwidth]{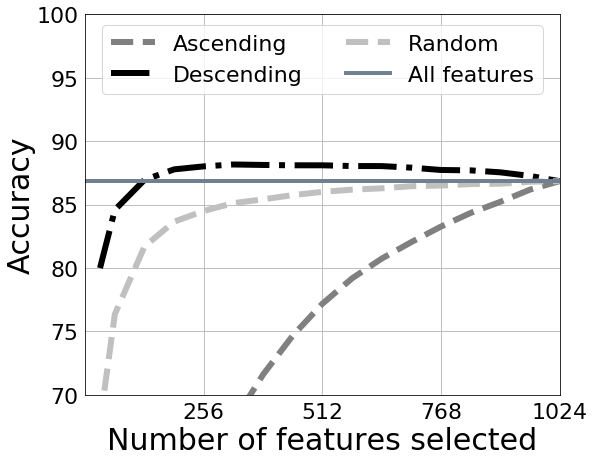}
      \end{minipage}
    \begin{minipage}{0.45\linewidth}
      \includegraphics[width = \linewidth]{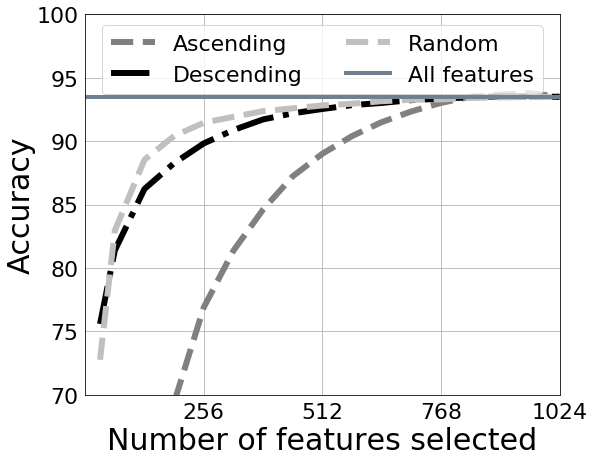}
    \end{minipage}
    \caption{\label{officeFigExp4}Mean accuracies over the 12 DA pairs with GoogleNet features using no adaptation (\textbf{left}) and the \textbf{OT3} adaptation algorithm (\textbf{right}). We note that the classification performances are better with all features when we apply an adaptation algorithm: 86.8\% without adaptation compared to 93.5\% with.}
    \end{figure}
    
    We saw in the previous experiment that our method works for different types of features with the best performances obtained using GoogleNet descriptors. However, these performances were achieved without applying any adaptation algorithm. To this end, we present in Figure \ref{officeFigExp4} the impact of using an adaptation algorithm that takes as input a reduced set of GoogleNet features returned by our method. Several important conclusions can be made based on these results. First, we notice that our algorithm does not improve the classification results compared to the performance of the \textbf{OT3} algorithm with a randomly selected subset of features. As explained in the introduction, \textbf{OT3} algorithm finds a new latent projection of source data in order to leverage the shift between the two domains. In this case, eliminating shifted features does not directly contribute to an improved classification performance as \textbf{OT3} algorithm can handle the reduction of shift between the two domains pretty well on its own. However, we can also observe the performance of \textbf{OT3} algorithm with a reduced "Ascending" set of features reaches its maximal value sooner than when no adaptation is performed. This is explained by the fact that the \textbf{OT3} algorithm successfully adapts the most shifted features. It is quite intuitive to assume that by selecting a subset of features, we decrease the computational complexity of the adaptation and classification algorithms that are used later. To support this claim, we present an additional study of the impact of reducing the number of features on both computational time and classification performance for several adaptation algorithms below. 
    
    \paragraph{Running time speed-up} For this experiment, we evaluated the gain in computational time of different adaptation algorithms as a function of the number of features selected by our method. To this end, we compared the ``no adaptation" setting with four state-of-the-art adaptation algorithms: \textbf{CORAL} \cite{sun2016return}, \textbf{SA} \cite{fernando2013unsupervised}, \textbf{TCA} \cite{pan2011domain} and \textbf{OT3} \cite{courty2014}. We fixed the subspace dimensions of \textbf{SA} and \textbf{TCA} to 80 (or to the number of feature selected when smaller than 80) while for \textbf{OT3} we set $\lambda=2$ and $\eta=1$. Even if from Table \ref{officeArrayEx3} we obtained the best performances with GoogleNet features, we select for this experiment the CaffeNet features to better see the computational gain because they have the largest dimensionality (4096).
    
    \setlength{\tabcolsep}{1mm}   
    \begin{table}[!t]
        \begin{tabular}{c c r | c r | c r | c r}
        \toprule
        Method & \multicolumn{2}{c}{$\searrow$512} & \multicolumn{2}{c}{$\searrow$1024} & \multicolumn{2}{c}{$\searrow$2048} & \multicolumn{2}{c}{4096}\\
        \midrule
        No adapt. & 79.2$\pm$2.2 &   0.00s & 79.9$\pm$2.3 &   0.00s & 80.0$\pm$2.2 &   0.00s & 74.4$\pm$3.0 &   0.00s\\
        \textbf{CORAL} & 80.5$\pm$1.8 & 110.43s & 80.8$\pm$1.9 & 587.69s & 80.4$\pm$1.7 & 3996.20s & 80.1$\pm$1.7 & 29930.39s\\
        \textbf{SA} & 81.8$\pm$2.0 &  13.25s & 82.5$\pm$1.8 &  32.09s & 82.9$\pm$1.7 &  66.71s & 83.0$\pm$1.7 & 169.71s\\
        \textbf{TCA} & 83.5$\pm$2.2 & 221.08s & 85.0$\pm$1.9 & 223.62s & 85.8$\pm$1.8 & 229.48s & 85.9$\pm$1.7 & 242.71s\\
        \textbf{OT3} & 84.2$\pm$2.4 &  19.50s & 86.7$\pm$1.9 &  31.76s & 88.8$\pm$1.5 &  54.07s & 88.8$\pm$1.4 &  97.47s\\
        \bottomrule
        \end{tabular}
    \caption{\label{officeArrayTime}Mean recognition accuracies in \%, standard deviation and sum of total computational time (over the 12 DA pairs and 19 iterations) in seconds for different adaptation algorithms using the CaffeNet features.}
    \end{table}
    
    The results of this evaluation are presented in Table \ref{officeArrayTime}. From these results, we see that by selecting 2048 out of 4096 most similar features, we are able to obtain slightly better classification performances for all adaptation methods compared to the case when all features are used. What's more, the computation time required by the algorithms greatly decrease. When only 512 features are used, an even more impressive speed up is obtained with a very slight drop in performance for the last three methods. These results confirm that our method is capable of finding subsets of similar features between source and target domains that can give comparable and sometimes even improved classification performances while decreasing considerably the computation time required for adaptation methods to converge.
    
%     \subsection{Experiments on prostate cancer mapping data set}
%     We now proceed to evaluate our method on a real-world data set collected for the problem of prostate cancer mapping.
    \subsection{Experiments on medical imaging data set}
    
    We now proceed to the evaluation of our method on a clinical data set of multiparametric magnetic resonance images (mp-MRI) collected to train a computer-aided diagnosis system for prostate cancer mapping \cite{niaf2012computer,DBLP:conf/icml/AljundiLPRL15}. This system learns a binary decision model in a multidimensional feature space based on training samples (voxels) from different classes of interest. This model is then used to generate cancer probability maps.
    
    \paragraph{Data description} The considered database consists of 90 mp-MRI exams acquired with different imaging protocols on two different scanners (49 patients on a 1.5T scanner and 41 on a 3T scanner), thus producing heterogeneous data sets.
    %     For this task, we possess two databases of annotated Magnetic Resonance Images (MRI) regrouping the exams of a total of 90 patients. One of these databases was acquired with a 1.5T scanner while the second one was obtained using a newer 3T scanner that produces MRI of higher resolution and of a different size. In this setting, we assume that the annotated data from the 1.5T scanner consisting of 49 patients represent the source domain while the one produced using the 3T scanner and containing 41 patients is the target domain. 
    %     Each individual voxel of the MRI of the two databases is described by a binary label (Cancer, Not Cancer) and a set of 95 features. 
     Each individual voxel is described by a binary label (Cancer, Non Cancer) and a set of 95 handcrafted features consisting of image descriptors, texture coefficients, gradients and other visual characteristics (more details in \cite{niaf2012computer}). Some of these 95 features have a clear shift between the two domains, as illustrated in Figure \ref{figureShift}. The number of available instances in both domains is shown in Table \ref{repartitionVoxels}. Our goal is to learn a classifier on annotated 1.5T voxels, representing the source domain, performing well on 3T voxels, considered as the target domain, without using labels from the latter one.    
    %     The number of available samples in both domains is shown in Table \ref{repartitionVoxels}. Our goal here would be to learn a classifier on annotated 1.5T voxels performing well on 3T voxels without using labels from the latter ones.
     \setlength{\tabcolsep}{1mm}
     \renewcommand*{\arraystretch}{1.2}
    \begin{table}[!t]
    \centering
    \begin{tabularx}{\textwidth}{>{\centering\arraybackslash} p{0.47\linewidth}>{\centering\arraybackslash} p{0.5\linewidth}}
        \begin{tabular}{l r r r}
            \toprule
            Class &   \#voxels 1.5T & \#voxels 3T \\
            \midrule
            Non cancer & 363,222 &  846,556 \\
            Cancer     &  56,126 &  140,840 \\
            \midrule
            Total      & 419,348 &  987,396 \\
            \bottomrule
        \end{tabular} 
        &
        \makebox[0.48\columnwidth][c]{
            \includegraphics[width=0.27\columnwidth, valign=m]{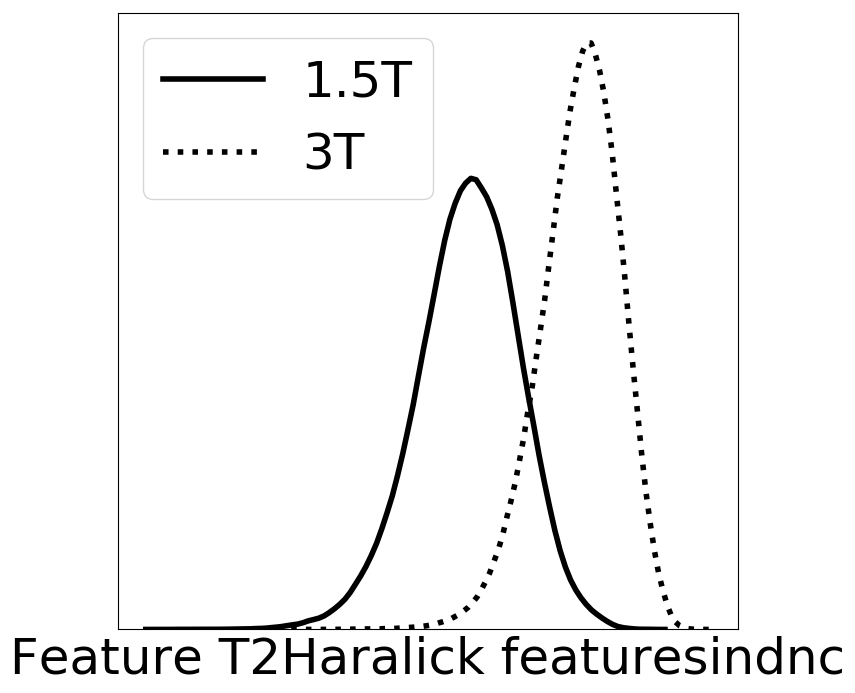}
            \includegraphics[width=0.26\columnwidth, valign=m]{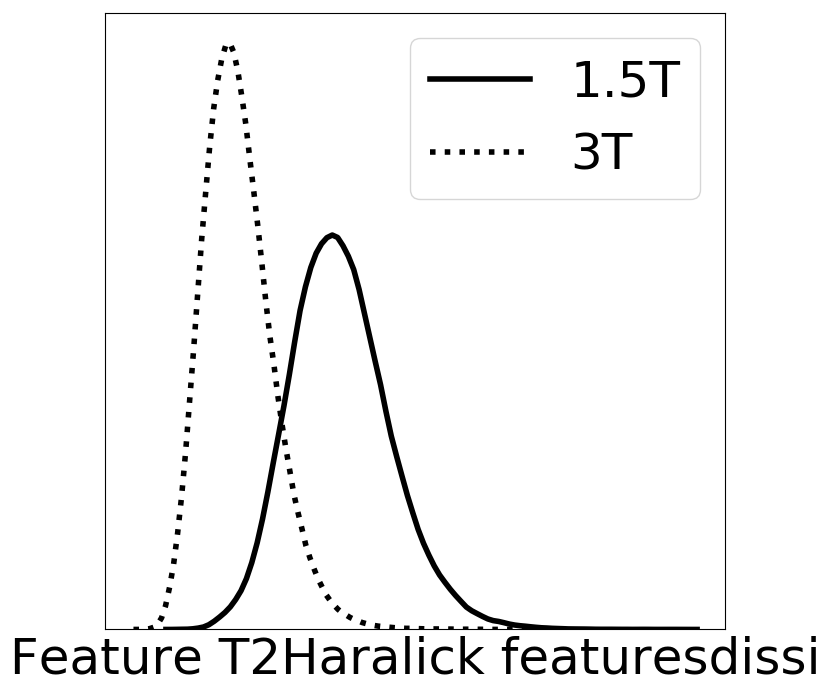}}
         \\ 
        \captionof{table}{Repartition of the MRI voxels between the Cancer and Non Cancer classes in source and target domains.}
        \label{repartitionVoxels}
        &
        \captionof{figure}{Example distribution of 2 features illustrating the shift between the source and target domains.}
        \label{figureShift}
    \end{tabularx}
    %\vspace{-0.5cm}
    \end{table} 

	\paragraph{Evaluation protocol} We first randomly sample a set $\bm{S}$ of 1500 voxels equiproportionally from the 49 1.5T exams and both classes of interest. Then, we use Algorithm \ref{algoSampleSelection} on $\bm{S}$ and on $\bm{T}$ as 20000 randomly sampled voxels from the 41 3T exams to obtain $\bm{T_u}$. This step is followed by the adaptation of $\bm{S}$ to $\bm{T_u}$, training a linear SVM on $\bm{S_a}$ and testing it on all voxels from the 3T target domain. 

    % This step was followed by adapting $\bm{S}$ to $\bm{T_u}$ and further a linear SVM from the source adapted data and testing it on all the voxels from the 3T database. As a performance measure, we use the area under the ROC Curve (AUC) which is a commonly used in the medical domain.
    We used the area under the ROC Curve (AUC) as the diagnostic performance measure. This is due to the fact that both the source and the target domains data exhibit an important class imbalance with $86\%$ of non-cancer voxels. In this case, the classification accuracy used in the previous experiments does not provide a truthful picture of the classifier's performance.  
    Our feature selection method is used as a standalone method and in combination with the \textbf{OT3} adaptation algorithm. As before, we repeat this process 20 times, and we report the mean AUC over the 20 iterations.
    
    \begin{figure}[!t]
    \centering
    \begin{minipage}{0.45\linewidth}
      \includegraphics[width=\columnwidth]{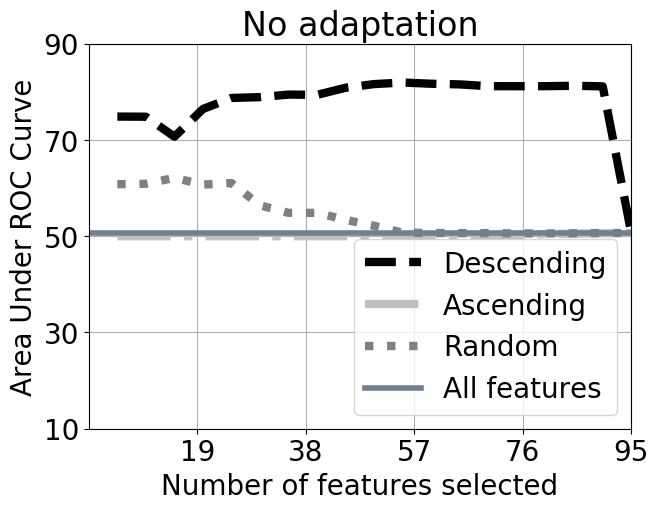}
      \end{minipage}
      \hfill
    \begin{minipage}{0.45\linewidth}
      \includegraphics[width = \linewidth]{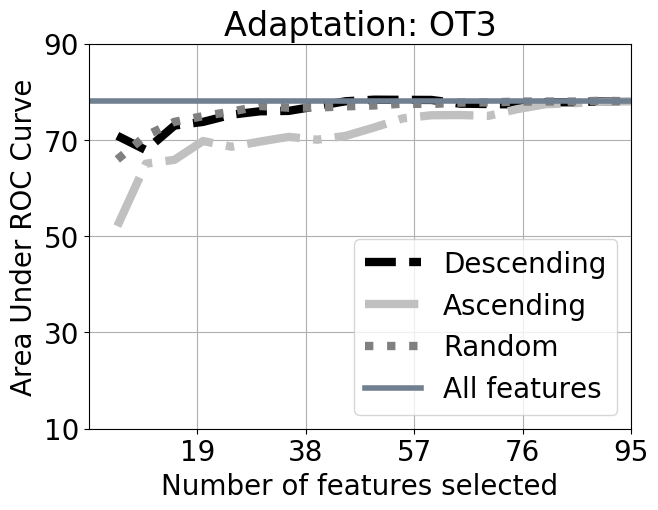}
    \end{minipage}
    \begin{minipage}{0.9\linewidth}
        \centering
        \includegraphics[width = \linewidth]{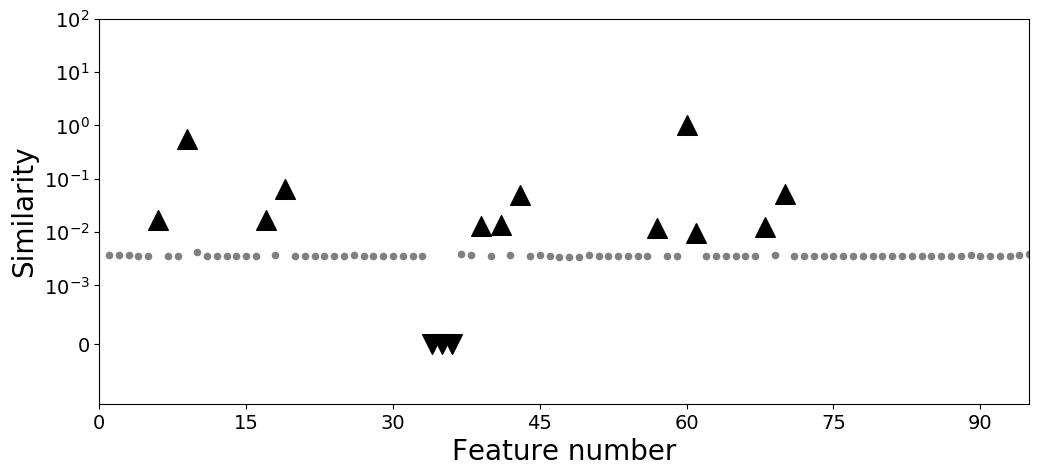}
    \end{minipage}
    \caption{\label{prostateFeatureSelection}Performance of our method on the clinical MRI database with no adaptation (top row, left) and using the \textbf{OT3} algorithm (top row, right). The log-scaled similarity of features across the two domains estimated by our algorithm is given in the bottom row. We observe that our method correctly identifies the three most shifted features that lead to an important drop in classifier's performance.}
    \end{figure}
    
    \paragraph{Obtained results}
    The results for this data set are shown in Figure \ref{prostateFeatureSelection}. When all the 95 features are used, we obtain an AUC of 50\% without adaptation, corresponding to the worst possible performance with no distinction between Cancer and Non cancer classes. By applying our feature selection algorithm (the ``Descending" curve) in a standalone manner, we are able to reach an AUC of 80\% with a significant drop in performance when the 3 more dissimilar features are added. On the other hand and similar to the Office/Caltech data set, using our feature selection algorithm before applying an adaptation algorithm reduces greatly the number of features needed to achieve comparable performance. This benefit presents an important computational gain when high-dimensional data sets are considered. Finally, we argued that one of the strengths of our method is its ability to identify the original features causing the shift between the source and target domains. To this end, we plot in Figure \ref{prostateFeatureSelection} the coupling values used to order features by their similarity across the two domains. From this Figure, we can see that our algorithm allows to identify the three most shifted features that lead to a significant performance drop observed previously. 
    
    \section{Conclusions and future perspectives}
    \label{sec:conclusions}
    In this paper, we presented a new feature selection method for domain adaptation based on optimal transport. Building upon a recent theoretical work on optimal transport in domain adaptation, we proposed a feature selection method that transports the empirical distribution of features in the source domain to that of the target one in order to obtain a coupling matrix representing their joint distribution. This coupling matrix is further used to identify the subset of features that remain unshifted across the two domains. We evaluated our method on both benchmark and real-world data sets and showed its efficiency in identifying the subset of features that successfully reduces the discrepancy between the two domains. Furthermore, we illustrated the usefulness of our method in reducing the computational time of several state-of-the-art methods that converge faster when taking as input a reduced set of features returned by our algorithm. 
    
    The possible future investigations that may follow up the presented work are many. First of all, we would like to combine our feature selection algorithm with a feature-transformation domain adaptation algorithm in a way such that the projection of data and the selection of features would be performed simultaneously. The potential interest of this joint approach would be to reduce the computational complexity of the adaptation methods and to improve their performance while maintaining the ease of interpretability of the obtained results. On the other hand, it would be also very interesting to extend the proposed framework to the general transfer learning scenario where the source and target domains tasks are not necessarily the same. In this case, the feature selection algorithm would have to take into account the discriminative power of each source feature in the target domain. Solving this problem in an unsupervised setting is a very challenging task that would require an efficient feature expressiveness measure to be introduced. We believe that this future perspective would be of a great interest in many real-world applications, notably the health-care one, where the manual labeling of the produced MRI scans represents an important bottleneck due to its highly time-consuming nature. 
    
    %References and End of Paper
    %These lines must be placed at the end of your paper
    \bibliography{ms}

\begin{thebibliography}{10}
\providecommand{\url}[1]{\texttt{#1}}
\providecommand{\urlprefix}{URL }
\providecommand{\doi}[1]{https://doi.org/#1}

\bibitem{DBLP:conf/icml/AljundiLPRL15}
Aljundi, R., Lehaire, J., Prost{-}Boucle, F., Rouvi{\`{e}}re, O., Lartizien,
  C.: Transfer learning for prostate cancer mapping based on multicentric {MR}
  imaging databases. In: Machine Learning Meets Medical Imaging workshop at
  ICML. pp. 74--82 (2015)

\bibitem{bay2006}
Bay, H., Tuytelaars, T., Van~Gool, L.: Surf: Speeded up robust features. In:
  ECCV (2006)

\bibitem{bendavidth}
Ben-David, S., Blitzer, J., Crammer, K., Kulesza, A., Pereira, F., Vaughan, J.:
  A theory of learning from different domains. Machine Learning  \textbf{79},
  151--175 (2010)

\bibitem{Ben-david07analysisof}
Ben-David, S., Blitzer, J., Crammer, K., Pereira, F.: Analysis of
  representations for domain adaptation. In: NIPS. pp. 137--144 (2007)

\bibitem{courty2014}
Courty, N., Flamary, R., Tuia, D.: Domain adaptation with regularized optimal
  transport. In: ECML/PKDD. pp. 274--289 (2014)

\bibitem{cuturi2013}
Cuturi, M.: Sinkhorn distances: Lightspeed computation of optimal transport.
  In: NIPS. pp. 2292--2300 (2013)

\bibitem{fernando2013unsupervised}
Fernando, B., Habrard, A., Sebban, M., Tuytelaars, T.: Unsupervised visual
  domain adaptation using subspace alignment. In: ICCV. pp. 2960--2967 (2013)

\bibitem{Gopalan:2011:DAO}
Gopalan, R., Li, R., Chellappa, R.: Domain adaptation for object recognition:
  An unsupervised approach. In: ICCV. pp. 999--1006 (2011)

\bibitem{gretton2006kernel}
Gretton, A., Borgwardt, K.M., Rasch, M.J., Sch\"{o}lkopf, B., Smola, A.: A
  kernel two-sample test. Journal of machine learning research  \textbf{13},
  723--773 (2012)

\bibitem{guyon2003introduction}
Guyon, I., Elisseeff, A.: An introduction to variable and feature selection.
  Journal of machine learning research  \textbf{3},  1157--1182 (2003)

\bibitem{jia2014}
Jia, Y., Shelhamer, E., Donahue, J., Karayev, S., Long, J., Girshick, R.,
  Guadarrama, S., Darrell, T.: Caffe: Convolutional architecture for fast
  feature embedding. In: International conference on Multimedia. pp. 675--678
  (2014)

\bibitem{knight2008}
Knight, P.A.: The sinkhorn--knopp algorithm: convergence and applications. SIAM
  Journal on Matrix Analysis and Applications  \textbf{30}(1),  261--275 (2008)

\bibitem{krizhevsky2012}
Krizhevsky, A., Sutskever, I., Hinton, G.: Imagenet classification with deep
  convolutional neural networks. In: NIPS. pp. 1097--1105 (2012)

\bibitem{li2016joint}
Li, J., Zhao, J., Lu, K.: Joint feature selection and structure preservation
  for domain adaptation. In: IJCAI. pp. 1697--1703 (2016)

\bibitem{niaf2012computer}
Niaf, E., Rouvi{\`e}re, O., M{\`e}ge-Lechevallier, F., Bratan, F., Lartizien,
  C.: Computer-aided diagnosis of prostate cancer in the peripheral zone using
  multiparametric mri. Physics in medicine and biology  \textbf{57}(12),
  3833--51 (2012)

\bibitem{pan2011domain}
Pan, S.J., Tsang, I.W., Kwok, J.T., Yang, Q.: Domain adaptation via transfer
  component analysis. IEEE Transactions on Neural Networks  \textbf{22}(2),
  199--210 (2011)

\bibitem{persello2016kernel}
Persello, C., Bruzzone, L.: Kernel-based domain-invariant feature selection in
  hyperspectral images for transfer learning. IEEE Transactions on Geoscience
  and Remote Sensing  \textbf{54}(5),  2615--2626 (2016)

\bibitem{DBLP:journals/corr/RedkoHS16}
Redko, I., Habrard, A., Sebban, M.: Theoretical analysis of domain adaptation
  with optimal transport. In: ECML/PKDD. pp. 737--753 (2016)

\bibitem{Saenko:2010:AVC}
Saenko, K., Kulis, B., Fritz, M., Darrell, T.: Adapting visual category models
  to new domains. In: ECCV. pp. 213--226 (2010)

\bibitem{sun2016return}
Sun, B., Feng, J., Saenko, K.: Return of frustratingly easy domain adaptation.
  In: AAAI. p.~8 (2016)

\bibitem{szegedy2015}
Szegedy, C., Liu, W., Jia, Y., Sermanet, P., Reed, S., Anguelov, D., Erhan, D.,
  Vanhoucke, V., Rabinovich, A.: Going deeper with convolutions. In: CVPR.
  pp.~1--9 (2015)

\bibitem{Talagrand95IHES}
Talagrand, M.: Concentration of measure and isoperimetric inequalities in
  product spaces. Publications Math\'ematiques de l'{I.H.E.S.}  \textbf{81},
  73--205 (1995)

\bibitem{uguroglu11}
Uguroglu, S., Carbonell, J.: Feature selection for transfer learning. In:
  ECML/PKDD. pp. 430--442 (2011)

\bibitem{Villani2008}
Villani, C.: Optimal transport: old and new, vol.~338. Springer Science \&
  Business Media (2008)

\bibitem{yin2017cross}
Yin, Z., Wang, Y., Liu, L., Zhang, W., Zhang, J.: Cross-subject eeg feature
  selection for emotion recognition using transfer recursive feature
  elimination. Frontiers in neurorobotics  \textbf{11} (2017)

\end{thebibliography}
    \bibliographystyle{splncs04}
 
\newpage
\appendix
\begin{center}\Large
Supplementary material: Feature Selection for Unsupervised Domain Adaptation using Optimal Transport
\end{center}
 \section{Comparison of Optimal Transport based methods}
\begin{figure}
    \centering
        \includegraphics[width=0.9\textwidth]{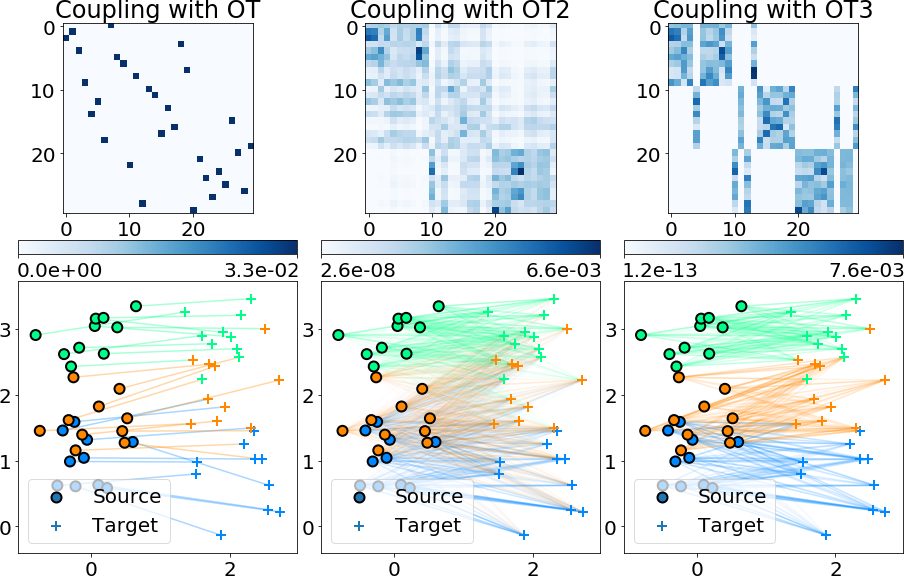}
        \caption{\label{comparisonOT}Comparison of the 3 variants of optimal transport: \textbf{OT} on the left, \textbf{OT2} in the middle ($\lambda=1$), \textbf{OT3} on the right ($\lambda=1$, $\eta=1$). First row shows $\gamma^*$ with highest coupling values seen as darkest blue. Second row shows the source and target points composed of 3 classes in 3 colors. The coupling between them are shown as segments.}
    \end{figure}

    We begin this Supplementary material by comparing in Figure \ref{comparisonOT} the three optimal transport algorithms introduced and used in the main paper. We see that the basic \textbf{OT} associates one target instance to one source instance while with \textbf{OT2} each source point's mass is divided and transported to its closest target points. By adding the class regularization \textbf{OT3}, we prevent to transport the mass of source instances of different classes to the same target instance.
    \newpage
 \section{Additional experiments}
    We now present a series of additional empirical evaluations that illustrate different important properties of the proposed algorithm. We start with a study of the impact that the algorithm used to find the shared representation for features can have on the obtained performance. 
    
    \subsection{Comparison of different samples selection strategies} As explained in the main paper, our method requires to select a set of examples that describe the features in the source and target domains before computing the similarity between them (the features). For our method, we propose to select these examples using the \textbf{OT} algorithm between the source and target samples. In Table \ref{officeArrayEx2}, we evaluate two other sample selection methods on the Office/Caltech data set: first one is based on the random selection of the examples while the second uses a 1-Nearest-Neighbor (1NN) algorithm instead of the \textbf{OT} method. The computation of the features' rank is then done in the same way as presented in the main article.
    
\begin{figure}[!ht]
        \centering
        \includegraphics[width=0.9\textwidth]{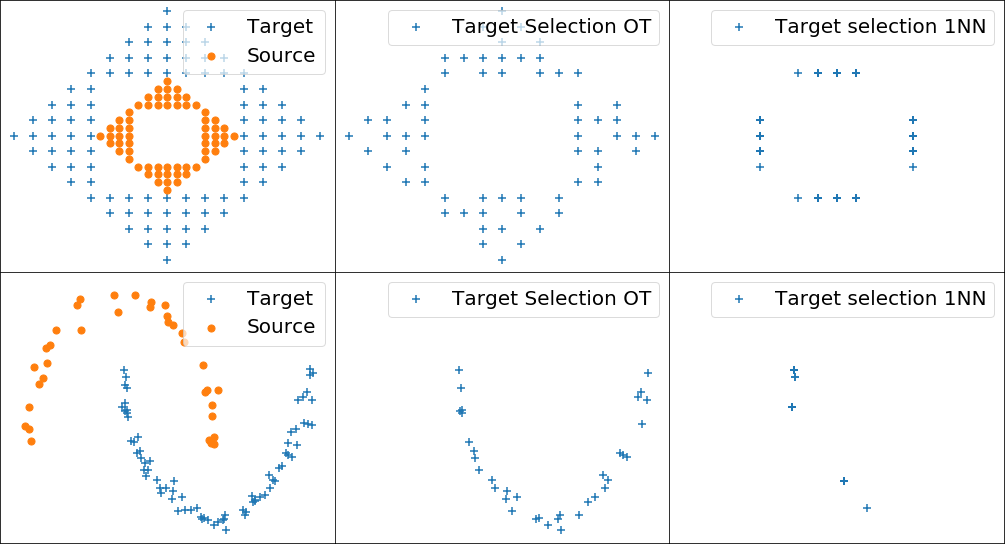}
        \caption{\label{figToy}Two toy examples where we generated a source and a target distribution (left) before using the sample selection procedure in the target domain using the \textbf{OT} algorithm (in the middle) and the 1NN selection (on the right).}
    \end{figure}

    \setlength{\tabcolsep}{6mm}
    \begin{table}[!ht]
        \centering
        \begin{tabular}{c c c c}
            \toprule        
            \#features & Random & \textbf{OT} & 1NN\\
            \midrule
            $\searrow$128 & 42.7$\pm$6.0 & \textbf{74.6$\pm$3.4} & 72.8$\pm$2.9\\
            $\nearrow$128 & 43.9$\pm$5.6 & 20.8$\pm$2.7 & 22.3$\pm$3.1\\
            \midrule
            $\searrow$512 & 68.1$\pm$4.5 & \textbf{79.3$\pm$2.6} & 79.1$\pm$2.7\\
            $\nearrow$512 & 60.8$\pm$5.3 & 27.1$\pm$3.3 & 27.6$\pm$3.4\\
            \midrule
            $\searrow$2048 & 75.9$\pm$3.2 & \textbf{80.1$\pm$2.2} & 79.6$\pm$2.7\\
            $\nearrow$2048 & 68.9$\pm$4.8 & 52.3$\pm$4.2 & 50.4$\pm$4.4\\
            \midrule
            \phantom{$\searrow$}4096 & 75.2$\pm$3.0 & 75.2$\pm$3.0 & 75.2$\pm$3.0\\
            \bottomrule
        \end{tabular}
        \caption{\label{officeArrayEx2}Mean accuracies over the 12 adaptation pairs without applying adaptation on CaffeNet features obtained using different sample selection methods. Our proposed selection method is \textbf{OT}.}
    \end{table}
    %With the 512 most similar features, we obtain for Random, OT and 1NN accuracies of respectively 68.1\%, 79.3\% and 79.1\%. For 128 features, we see similar differences between the three methods with accuracies of respectively 42.7, 74.6 and 72.8. On the other side by selecting first the features the most dissimilar, with 512 features we obtain accuracies 60.8, 27.1 and 27.6, and with 128 features accuracies 43.9, 20.8 and 22.3. The random selection do not present large differences in performances by either taking the most similar or dissimilar features. With our proposed method however, there is a large differences: with 128 features selected, we obtain a significant difference of accuracy of 53.8. This prove the interest of our method to not only find similar features, but also dissimilar. We see that the values of the selection using the 1NN algorithm give similar performances to our proposed method OT. 
    From this table, we can see that random selection of instances gives poor results for different numbers of features considered in our study. On the other hand, we observe that both 
\textbf{OT} and 1NN algorithm provide close performances in identifying both similar and dissimilar features with a slight superiority of the optimal transport based method. In order to separate the two, we demonstrate in Figure \ref{figToy} the pitfalls of 1NN based selection that can occur when the vast majority of source points are associated with a handful of target instances. 

    %\paragraph{Toy examples for sample selection}
    %We saw previously that the sample selection prior to our feature ranking method is important: selecting randomly the samples do not work well. But we also saw that the selection with OT and 1NN give similar performances. We propose in 

	From this figure, we can see that for the two considered toy data sets, the selection of target instances based on the 1NN algorithm leads to a distribution that does not reflect the true distribution of the target data. If we would have selected points in the target domain randomly, we would still have the same target distribution, but as we have shown previously in Table \ref{officeArrayEx2}, the random selection gives worse classification performances. The proposed strategy for the selection of target samples through the \textbf{OT} algorithm allows to obtain both good classification performances and to preserve the target data distribution. On the other hand, the computation of the \textbf{OT} has a squared space complexity compared to the linear complexity of the 1NN selection. Consequently, we note that the use of the sample selection with the 1NN algorithm can present a good alternative for large-scale machine learning problems.

    \subsection{Classification results for SURF and GoogleNet descriptors}
\begin{figure}[!th]
        \centering
        \includegraphics[width=0.49\textwidth]{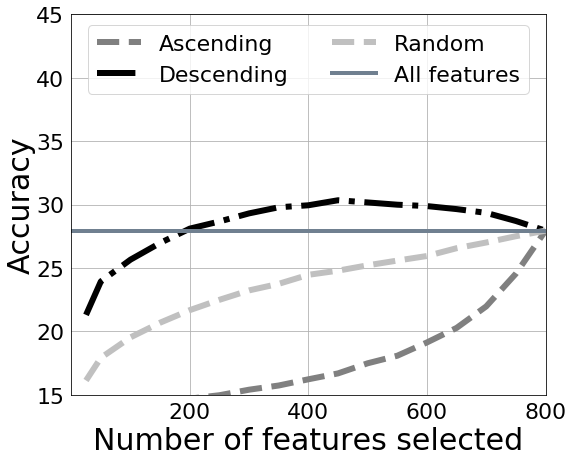}
        \includegraphics[width=0.49\textwidth]{images/NA_GoogleNet1024.png}
        \caption{Mean accuracy results from Table \ref{officeArrayNoAdapt3Features} for SURF (\textbf{left}) and GoogleNet (\textbf{right}) features.}
        \label{fig:surf_google}
    \end{figure}

    In the main paper, we only presented the classification results for the CaffeNet features. To this end, Table \ref{officeArrayNoAdapt3Features} and Figure \ref{fig:surf_google} provide the same results for two other types of descriptors considered in our work that are SURF and GoogleNet features.
    
    We observe the same behavior as for the Caffe features where our method gives better or almost identical performances on almost all domain adaptation pairs with significantly less features used. Thus, they confirm our claim about the efficiency of the proposed method for domain adaptation.  

     \setlength{\tabcolsep}{3mm}
    \begin{table}
    \begin{tabular}{c @{\hspace{-0.5\tabcolsep}}|@{\hspace{-0.5\tabcolsep}} c @{\hspace{-0.5\tabcolsep}}|@{\hspace{-0.5\tabcolsep}} c}
        \toprule
        & SURF features & GoogleNet features\\
        \begin{tabular}{c}
            DA pairs\\
            \midrule
            A$\rightarrow$C\\
            A$\rightarrow$D\\
            A$\rightarrow$W\\
            C$\rightarrow$A\\
            C$\rightarrow$D\\
            C$\rightarrow$W\\
            D$\rightarrow$A\\
            D$\rightarrow$C\\
            D$\rightarrow$W\\
            W$\rightarrow$A\\
            W$\rightarrow$C\\
            W$\rightarrow$D\\
            \midrule
            Mean\\
            \bottomrule
        \end{tabular}
        &
        \begin{tabular}{c c c}
            $\searrow$400 & $\nearrow$400 & 800\\
            \midrule
            \textbf{25.4$\pm$2.4} & 15.4$\pm$1.5 & 23.1$\pm$1.6\\
            \textbf{24.5$\pm$2.9} & 16.2$\pm$3.0 & 21.9$\pm$2.4\\
            \textbf{27.5$\pm$2.2} & 16.2$\pm$2.6 & 26.0$\pm$2.1\\
            \textbf{24.8$\pm$1.4} & 14.1$\pm$2.2 & 21.2$\pm$2.4\\
            \textbf{25.5$\pm$3.7} & 15.5$\pm$2.8 & 22.8$\pm$3.6\\
            \textbf{23.3$\pm$3.0} & 13.9$\pm$2.2 & 20.6$\pm$3.5\\
            25.7$\pm$2.0 & 15.8$\pm$2.9 & \textbf{26.7$\pm$1.7}\\
            23.8$\pm$1.9 & 16.0$\pm$2.1 & \textbf{24.8$\pm$1.5}\\
            \textbf{53.6$\pm$3.5} & 22.1$\pm$3.4 & 53.3$\pm$2.7\\
            \textbf{23.7$\pm$1.9} & 15.6$\pm$1.9 & 23.1$\pm$1.5\\
            18.1$\pm$1.7 & 12.0$\pm$1.6 & \textbf{19.5$\pm$1.0}\\
            \textbf{63.4$\pm$3.6} & 21.7$\pm$3.4 & 52.4$\pm$2.6\\
            \midrule
            \textbf{29.9$\pm$2.5} & 16.2$\pm$2.5 & 27.9$\pm$2.2\\
            \bottomrule
        \end{tabular}
        &
        \begin{tabular}{c c c}
        $\searrow$256 & $\nearrow$256 & 1024\\
        \midrule
        \textbf{85.7$\pm$1.2} & 64.7$\pm$2.4 & 84.6$\pm$1.1\\
        86.7$\pm$2.4 & 68.6$\pm$4.9 & \textbf{88.4$\pm$2.5}\\
        \textbf{85.4$\pm$3.1} & 51.8$\pm$5.9 & 83.5$\pm$2.8\\
        90.4$\pm$1.2 & 74.5$\pm$3.4 & \textbf{90.6$\pm$1.7}\\
        88.3$\pm$2.7 & 68.5$\pm$4.3 & \textbf{88.6$\pm$2.7}\\
        \textbf{86.2$\pm$2.7} & 54.3$\pm$4.6 & 83.3$\pm$2.4\\
        \textbf{84.2$\pm$2.0} & 46.4$\pm$4.2 & 82.3$\pm$1.6\\
        \textbf{80.5$\pm$1.7} & 46.9$\pm$2.8 & 77.8$\pm$2.5\\
        96.5$\pm$1.1 & 81.5$\pm$3.4 & \textbf{97.4$\pm$0.8}\\
        \textbf{89.7$\pm$0.8} & 55.8$\pm$2.5 & 87.0$\pm$1.2\\
        \textbf{83.7$\pm$1.1} & 49.9$\pm$2.5 & 79.4$\pm$1.2\\
        98.9$\pm$0.8 & 93.4$\pm$1.8 & \textbf{99.2$\pm$0.5}\\
        \midrule
        \textbf{88.0$\pm$1.7} & 63.0$\pm$3.5 & 86.8$\pm$1.8\\
        \bottomrule
        \end{tabular}
        
    \end{tabular}
    
    \caption{\label{officeArrayNoAdapt3Features}The arrays give the recognition accuracies in \% and standard deviation with no adaptation for SURF and GoogleNet features.}
    \vspace{-3cm}
\end{table}

\end{document}